\documentclass[journal]{IEEEtran}

\usepackage[utf8]{inputenc}
\usepackage[T1]{fontenc}
\usepackage{graphicx} 
\usepackage{amsmath}
\usepackage{amsfonts}
\usepackage{caption}
\usepackage{float} 
\usepackage{cite}
\usepackage{subcaption}
\usepackage{color}
\usepackage{lineno}
\usepackage{amssymb}
\usepackage{array}
\usepackage{booktabs}
\usepackage{multirow}
\usepackage[justification=centering]{caption}

\ifCLASSINFOpdf
  
\else
  
\fi

\begin{document}

\title{\LARGE \bf
    SpikeSMOKE: Spiking Neural Networks for Monocular 3D Object Detection with Cross-Scale Gated Coding

\author{Xuemei~Chen,~Huamin~Wang,~\IEEEmembership{Member,~IEEE,}~Jing~Peng,~Hangchi~Shen,~Shukai~Duan,~\IEEEmembership{Member,~IEEE,}~Shiping~Wen,~\IEEEmembership{Senior Member,~IEEE,}and~Tingwen~Huang,~\IEEEmembership{Fellow,~IEEE}

\thanks{© 2026 IEEE. Personal use of this material is permitted.
Permission from IEEE must be obtained for all other uses, in any current or future media,
including reprinting/republishing this material for advertising or promotional purposes,
creating new collective works, for resale or redistribution to servers or lists,
or reuse of any copyrighted component of this work in other works.

DOI: 10.1109/TETCI.2026.3670672}

}
}

\markboth{}%
{Shell \MakeLowercase{\textit{et al. }}: Bare Demo of IEEEtran. cls for IEEE Journals}

\maketitle

\begin{abstract}
With the wide application of 3D object detection in some fields such as autonomous driving, its energy consumption is constantly increasing, making the research on low-power consumption alternatives a key research area. The spiking neural networks (SNNs),  possessing low-power consumption characteristics, offer a novel solution for this research. Consequently, we apply SNNs to monocular 3D object detection and propose the SpikeSMOKE architecture, which represents a new attempt at low-power monocular 3D object detection. It's well known that  the discrete signals of SNNs can lead to information loss compared to artificial neural networks (ANNs), which limits their feature representation capabilities. To solve this problem, inspired by the synaptic filtering mechanism of biological neurons, we propose a new Cross-Scale Gating Coding Mechanism (CSGC), which can enhance feature representation by combining cross-scale fusion of attentional methods and gated filtering mechanisms. In addition, to reduce the computation and accelerate training, we present a novel light-weight residual block that can maintain spiking computing paradigm and the highest possible detection performance. Our method is effective on the KITTI, NuScenes-mini and CIFAR10/100 datasets.  Compared to the baseline SpikeSMOKE under the 3D Object Detection, the proposed SpikeSMOKE with CSGC can achieve 11.78 (+2.82, Easy), 10.69 (+3.2, Moderate), and 10.48 (+3.17, Hard) on the KITTI autonomous driving dataset by $AP|_{R_{11}}$ at 0.7 IoU threshold, respectively. It is worth noting that the results of SpikeSMOKE can significantly reduce energy consumption compared with the results of SMOKE. And SpikeSMOKE-L (lightweight) can further reduce the amount of parameters by 3 times and computation by 10 times compared to SMOKE.

\end{abstract}
\begin{IEEEkeywords}
    Spiking Neural Networks, Monocular 3D Object Detection, Gating Coding, Lightweighting. 
\end{IEEEkeywords}

\IEEEpeerreviewmaketitle

\section{Introduction}

\IEEEPARstart{A}{lthough} ANNs have shown excellent performance in various fields including computer vision (CV)\cite{16}, speech recognition (SR)\cite{17}, natural language processing (NLP)\cite{18}, etc, and have made great contributions to the development of artificial intelligence, they are facing with skyrocketing problems of energy consumption, with the increasingly complex models and progressively large volumes of data\cite{023}. Especially in 3D object detection, the maximum computational complexity observed in existing algorithms has already reached 50 GMAC\cite{022}. This not only increases hardware costs but also limits their applications in mobile devices and edge computing scenarios. To solve this problem, some researchers are turning to brain-inspired SNNs, because they have excellent characteristics of event-driven computation\cite{07}, biologically interpretable\cite{48}, and asynchronous temporal processing. These features enable SNNs to significantly reduce energy consumption and improve computational efficiency when dealing with sparse data and temporal dynamic signals.

At present, there has been a great deal of excellent research on SNNs applied to computer vision (CV), natural language processing (NLP) and other fields\cite{029}\cite{01}\cite{036}\cite{03}\cite{04}\cite{46}. In terms of the evolution of network architecture, the early Spiking ResNet\cite{08} laid the foundation for event-driven network architectures. Subsequently, EMS-ResNet\cite{06}   optimized this foundation, and MS-ResNet\cite{46}  further enhanced the network's ability to process complex data, bringing new ideas to the field of computer vision. In the realm of natural language processing, Spikformer\cite{09}, which is the initial utilization of Transformer in SNNs,  opened a new path for sequence modeling. The subsequent Spike-Driven Transformer v1/v2\cite{010}\cite{011}  continuously optimized, which not only offers robust sequence modeling capabilities, but also improves the efficiency of parallel processing of intricate language tasks. For object detection, current research focuses on 2D object detection using 2D image data or 3D object detection using 3D data. For example, in  Spiking-PointNet \cite{1}, point clouds—which represent 3D spatial information—were utilized for the purpose of detecting 3D objects, marking the first successful application of point clouds within SNNs. In aspect of 2D image data, SpikeYOLO\cite{3} has put forward a spike-driven object detection framework, which is capable of effectively extracting target features from 2D images and carrying out target localization and classification. However, SNNs still face some challenges in the field of object detection, especially in monocular 3D object detection. Since monocular 3D object detection requires processing additional depth information, traditional 2D object detection methods cannot be directly applied to 3D object detection. Up to now, there has been no mature exploration of monocular 3D object detection via SNNs, and the low-power characteristic of SNNs has not been utilized in this task.

Monocular 3D object detection combines 2D image information and depth estimation to achieve detection and localization of 3D objects, which has been widely used in the field of autonomous driving\cite{012}\cite{014}\cite{015}\cite{016}. Currently, there are many excellent models in the field of ANNs. For example, MonoWAD \cite{12} introduced an innovative weather-adaptive diffusion mechanism, which can effectively deal with changes in image features under different weather conditions, thereby enhancing the robustness and accuracy of monocular 3D object detection. LabelDistill\cite{14} presented a label-guided cross-modal knowledge distillation method, which uses data from other modalities to assist in 3D object detection from 2D images.  
Mono3DVG\cite{15}  offered a fresh perspective by detecting 3D objects through detailed descriptions of visual appearance, and is able to accurately identify and locate objects with complex appearances.  MonoCD\cite{24}, a network architecture based on auxiliary information, further optimized the performance of monocular 3D object detection. 
However,  when applying monocular 3D object detection to resource-constrained edge devices like autonomous driving\cite{017},  its practical deployment will be restricted. These edge devices typically have limited computing resources and energy supplies, while existing ANN models often require high computational power and energy consumption to implement complex detection algorithms. Therefore, it is very important to implement a network model with low-power characteristics for monocular 3D object detection. Against this backdrop, brain-inspired SNNs models have attracted attention for their unique low-power characteristics, which give them potential advantages in handling monocular 3D object detection tasks and offer new ideas for solving this problem.

Based on the above analysis, we apply brain-inspired Spiking Neural Networks (SNNs) with low-power characteristics to monocular 3D object detection to construct a novel architecture SpikeSMOKE. In order to shrink the gap between the discrete signals of SNNs and the continuous signals of ANNs for feature expressiveness, we propose a cross-scale gated coding mechanism (CSGC) inspired by the filtering mechanism of biological neuronal synapses, which can reduce information loss and improve the detection performance of the model by fusing multi-scale attention. To further reduce the model's power consumption while maintaining the spike computation paradigm, we design innovative light-weight residual blocks that can significantly reduce the computational load and energy consumption while maintaining the network's depth and complexity. To verify the effectiveness of the proposed method in this paper, we conducted a large number of experiments, which cover the widely used KITTI and NuScenes-mini datasets in the field of autonomous driving, as well as the CIFAR10 and CIFAR100 datasets in the field of computer vision. The contributions of this paper can be summarized as follows.

\quad (1) To enhance the energy efficiency of monocular 3D object detection, we propose a spiking architecture SpikeSMOKE based on SMOKE leveraging the low power consumption feature of brain-like SNNs.

\quad (2) To improve the information representation capability of SpikeSMOKE, we design a novel parallel cross-scale gated coding mechanism CSGC based on attention. Additionally, we propose a light-weight residual block integrated into the SpikeSMOKE model.

\quad (3) Compared to the baseline SpikeSMOKE, our enhanced model with CSGC achieves superior performance on the KITTI dataset. We further validate the generalizability of the proposed CSGC encoding strategy through experiments on the NuScenes-mini and CIFAR-10/100 datasets. Meanwhile, compared with SMOKE, SpikeSMOKE significantly reduces energy consumption in 3D object detection.

This paper is organized as follows. In section \ref{sec:related_work}, we describe the kinetic formulation of LIF for spike neurons, the model architecture for monocular 3D object detection, the  coding mechanism, and  lightweighting methods. In section \ref{sec:methods}, we describe the macroscopic architecture of SpikeSMOKE, the detailed implementation of cross-scale gated coding (CSGC) and lightweighted residual structures. In the \ref{sec:experiments} section, we present the experimental results of our methods and the results of ablation experiments. In the \ref{sec:conclusion} section, we summarize the paper.

\section{Related works}
\label{sec:related_work}

\subsection{Spike Neuron}

Spike neurons simulate the behavior of biological neurons transmitting signal sequences through synapses\cite{031}, transmitting temporal signals and spatial information through the membrane voltage U. This mechanism enables spike neurons to efficiently process dynamic information while maintaining low energy consumption, which is particularly significant when dealing with sparse data and time-series tasks. 
The commonly used neurons in spiking neural networks consist of integrate-and-fire (IF)\cite{019} neurons and leaky integrate-and-fire (LIF) \cite{018}\cite{034}neurons. These neuron models simulate the signal processing of biological neurons through different mechanisms\cite{035}. 
We used LIF neurons and described their neuronal dynamics formulation as follows:

\begin{equation}\label{eqLIF}
\begin{aligned}
&U_{i}^{l}[t]=H_{i}^{l}[t-1]+\sum_{j}W_{ij}S_{j}^{l}[t]\\
&S_{i}^{l}[t]=Hea(U_{i}^{l}[t]-Uth)=\begin{cases}1, U_{i}^{l}[t] \geq U_{th} \\0, U_{i}^{l}[t]<U_{th}\end{cases}\\
&H_{i}^{l}[t]=\tau U_{i}^{l}[t](1-S_{i}^{l}[t])+U_{reset}S_{i}^{l}[t]
\end{aligned}
\end{equation}

$U_{i}^{l}[t]$ denotes the membrane voltage of neuron i in layer l, which is the core variable of the neuron's state and reflects the excitation level of the neuron at time t. $W_{ij}$ indicates the connection weights of layer l and layer l-1, which determine how the spikes of neurons in the previous layer affect the membrane voltage of neurons in the current layer. $S_{j}^{l}[t]$  represents the spike of the jth neuron in layer l-1, and spikes are the fundamental units of information transmission between neurons. The Heaviside function, denoted as $Hea$, is used to determine whether the neuron has reached the firing threshold.
At discrete time step T, when the membrane voltage of the neuron is greater than the spike neuron threshold, i.e., $U_{i}^{l}[t]$$\geq$$U_{th}$, the neuron emit a spike\cite{11}. This process mimics the behavior of biological neurons firing action potentials when they reach a certain level of excitation. $\tau$ denotes the decay coefficient, which controls the rate of decay of the membrane voltage, reflecting the "memory" characteristic of the neuron\cite{030}. After issuing spike 1, the membrane voltage is usually reset to 0.

\subsection{Monocular 3D Object Detection}
Monocular 3D object detection relies solely on the 2D image information obtained from a monocular camera to  detect and locate 3D objects in a scene. However, the lack of depth information in monocular images makes it highly challenging to recover 3D information from 2D images. To address this challenge, researchers have proposed a variety of methods for monocular 3D object detection, which can generally be divided into three categories: two-stage methods, single-stage anchor-based methods, and single-stage anchor-free methods. Two-stage methods, such as MonoLSS\cite{27}, typically first generate a series of candidate regions and then perform fine classification and regression on these regions to determine the final 3D object bounding boxes. This method is prone to introducing 2D noise\cite{2}, which to some extent affects the accuracy and reliability of detection. Single-stage anchor-based methods, such as GrooMeD-NMS\cite{21}, define a series of anchor points on the image in advance and then perform classification and regression on these anchor points to directly generate 3D bounding boxes. These methods usually require complex data preprocessing and non-maximum suppression\cite{020}\cite{021}. Compared with the first two categories of methods, single-stage anchor-free methods (SMOKE\cite{2}, MonoCD \cite{24}, MonoDGP \cite{23}) have the significant advantages of a simple model and high computational efficiency, making them more suitable for real-time operation on resource-constrained edge devices.

In this paper, we use the SMOKE \cite{2} architecture of Single-stage anchor-free methods as a base architecture, which both ignores the redundancy associated with 2D detection frameworks and does not require additional data. The SMOKE architecture continues the keypoint detection of centernet \cite{43}, which discards the traditional 2D detection module and retains only the 3D detection part, and improves the parameter convergence and detection accuracy by multi-step disentanglement. Specifically, it decomposes the attribute regression of the 3D bounding box into multiple steps, optimizing each parameter progressively. The 3D bounding box is obtained by predicting the object's 3D projection center on the image plane along with attribute variables.  The attribute regression of the 3D bounding box is decoupled into an 8-tuple $(\alpha_x, \alpha_y, \alpha_z, \alpha_l, \alpha_w, \alpha_h, sin\beta, cos\beta)$. These parameters are optimized by a loss function, and then transformed into the actual parameters for constructing the 3D bounding box, including the object's center position $(x,y,z)$, dimensions $(w,h,l)$, and orientation angle $(\theta)$.

\subsection{Coding Mechanism}
In a spiking neural network, data encoding is a crucial step for efficient information processing. Initially, input data must be transformed into spike sequences for further processing by neurons, a process that involves selecting from several encoding methods. Rating coding \cite{39} encodes information based on the average rate of neuron firing, the higher the firing rate of neurons, the greater the signal strength or information content they represent. Its advantage lies in its simple implementation and similarity to the behavior of biological neurons. However, it requires a longer time window to accurately estimate the firing rate, which may lead to information delay. Phase coding \cite{38} expresses information by the temporal position of spikes relative to a reference event. For instance, in a periodic stimulus, the phase of a spike can precisely indicate specific signal features. Temporal coding \cite{7} represents information based on the time intervals and order of spikes, effectively utilizing the temporal characteristics of spikes to handle dynamic and sequential data. Frequency coding involves the frequency at which neurons emit spikes over a specific time period. It represents information through the repetition rate of spikes and emphasizes short-term frequency changes rather than long-term average firing rates. In addition to this, there are various complex coding methods (\cite{8}\cite{40} ), and choosing an appropriate coding method is crucial for the representation performance of the model. In order to reduce the gap between the discrete and continuous signals in the spiking neural network, we propose a novel gated coding method in conjunction with the cross-scale fusion unit. This encoding method introduces a gating mechanism that more effectively controls the flow and processing of information.

\begin{figure*}[htbp]
    \centering
    \makebox[\textwidth][c]{\includegraphics[width=1\textwidth]{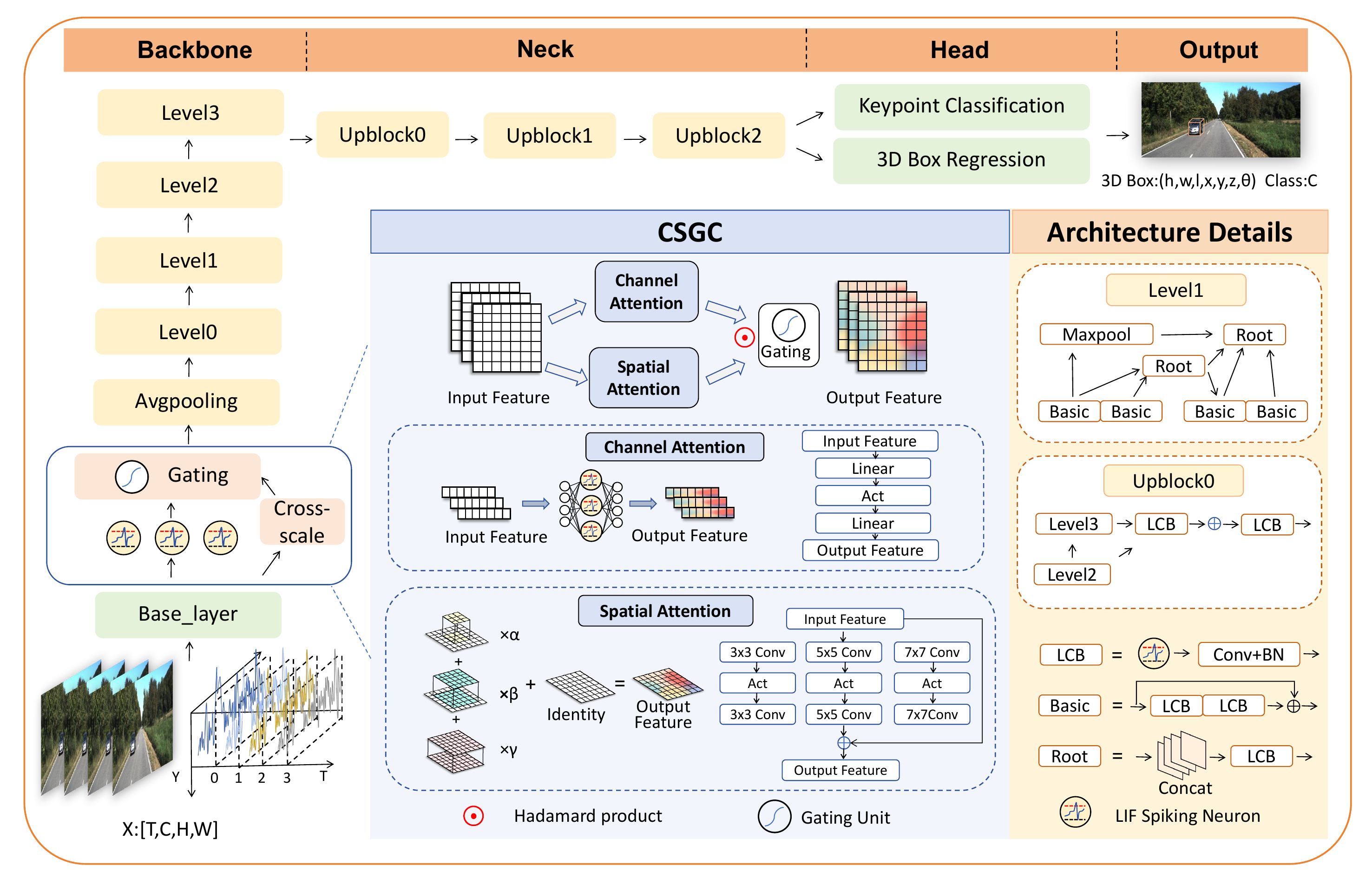}}%
    \caption{\small \small SpikeSMOKE overall network architecture. The input is an image and the output is the 3D bounding box of the object detection and the object category. A CSGC coding mechanism is introduced, based on a cross-scale attention fusion, to reduce the performance loss due to data transformation and enhance the representation of complex features. }
    \label{fig2}
\end{figure*}

\subsection{Light-Weight Model}
Light-weight model refers to reducing the number of parameters and computation and compressing the model training time by removing or transforming part of the model architecture and connections\cite{032}\cite{033}. There are several common methods for lightweighting models in neural networks, a) Pruning: Chen et al \cite{29} proposed an SNN pruning regeneration mechanism based on neuron criticality to remove unnecessary connections or neurons in the network. This mechanism can identify and remove connections or neurons that contribute less to the model's performance, thereby simplifying the network structure and reducing unnecessary computations. SSNN \cite{28} designed sparse RSNN by pruning randomly initialized model. 
b) Model quantization: Model quantization reduces the model's storage requirements and computational load significantly by converting the model's weights and activations from floating-point numbers to low-bit representations, such as binary or ternary quantization. Spiking-Diffusion \cite{30} proposed a diffusion model based on vector quantization discretization. c) Model distillation:  TSSD \cite{32} proposed a spatio-temporal distillation method, which transfers the knowledge from a complex large model (the teacher model) to a simpler small model (the student model). This enables the student model to inherit the performance of the teacher model while maintaining low computational complexity.
d) Depth-wise separable convolutions: Depth-wise separable convolutions decompose standard convolutions into two steps: depth-wise convolutions and point-wise convolutions, significantly reducing the computational load of convolution operations. The first proposed in MobileNets \cite{9}, and later extended and applied widely in the field of ANN ( \cite{33},  \cite{34}). There has been a great deal of experimental or theoretical evidence for the effectiveness of depth-wise separable convolutions, and our proposed lightweight residual unit originates from this idea, aiming to further optimize the model's computational efficiency and performance.

\section{Methods}
\label{sec:methods}

\subsection{SpikeSMOKE}
The SpikeSMOKE architecture is shown in Figure \ref{fig2}. The design of this architecture is based on the single-stage object detection network SMOKE, which has the advantages of ignoring the redundancy of 2D detection frameworks and not requiring additional data. We choose this architecture because it can simplify the complexity of the model while maintaining efficient detection performance.

Backbone: The  architecture DLA34 can take advantage of fusing feature maps of different scales through the depth convergence feature. To fully transform the model into a SNNs-based architecture, we introduced the spiking rate of spiking neurons to replace the ReLU activation function in DLA34, which is very crucial. Meanwhile, to enhance the processing capability of spike signals, we have proposed the CSGC module in backbone, including channel attention mechanism and spatial attention mechanism. A detailed introduction to CSGC is in Section Methods B.

Neck: The design of the neck aims to further optimize the transmission and processing of features, ensuring that the model is fully driven by spikes. In the DLAup structure, we ensure that a spiking neuron is used before each convolution operation to convert the continuous signals into discrete spike signals, thereby maintaining the spike-driven characteristics.  Meanwhile, deformable convolutions were employed to further enhance feature extraction.

Head: This component consists two branches. One is keypoint classification used by heatmap, which is mainly responsible for analyzing the input feature maps to identify the locations of object keypoints and generate heatmaps to represent the distribution of these keypoints. The other is 3D bounding box regression, which is in charge of predicting the 3D bounding box parameters of each detected object based on the information in the feature maps, including the center coordinates, size, and rotation angles of the bounding boxes.  This two branches can process the feature maps from the Neck network to obtain the 3D object detection results.

\begin{figure}[htbp]
    \centering
    \includegraphics[width=0.8\columnwidth]{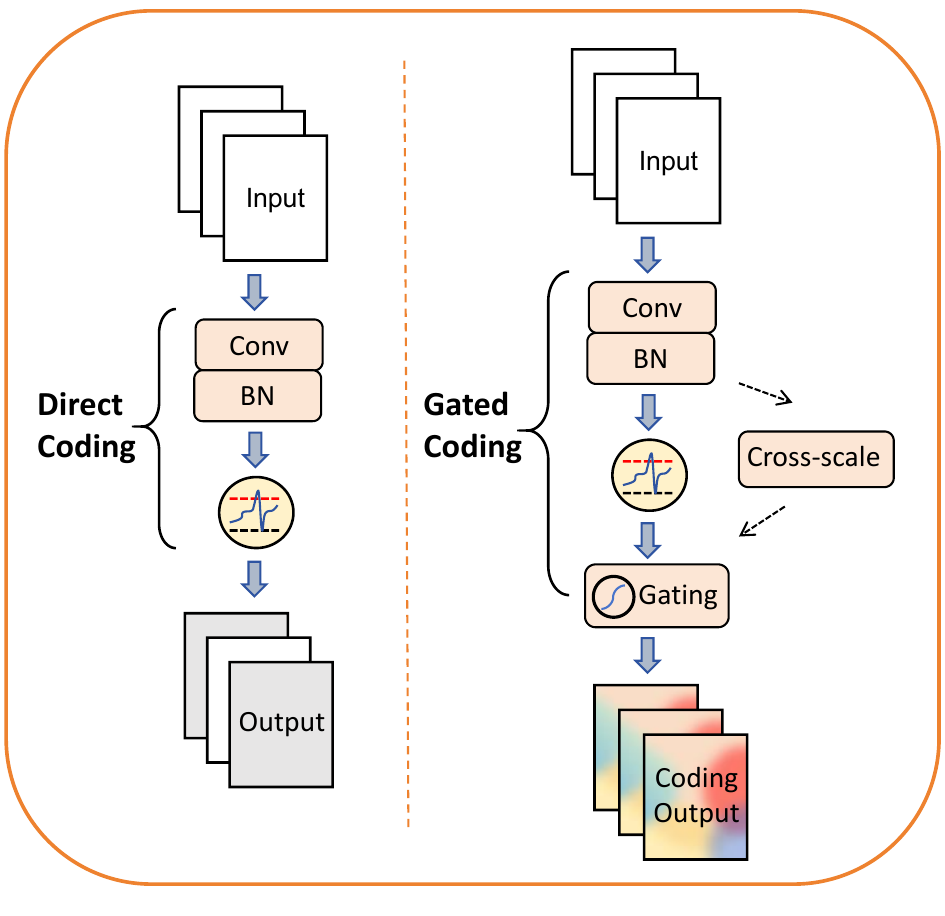} 
    \captionsetup{justification=raggedright, singlelinecheck=false} 
    \caption{\small \small Gated coding is different from direct coding. The proposed gated coding mechanism utilizes a cross-scale fusion module that incorporates a gating unit (CSGC) to mimic the filtering mechanism of biological neuronal synapses to produce a filtering effect on the input feature map. }
    \label{fig1}
\end{figure}

\subsection{Cross-Scale Gated Coding (CSGC)}
When converting ANN to SNN, the fundamental contradiction of sparsity and discreteness of SNN is faced, which weakens the modeling ability of continuous spatial structures. Especially in monocular 3D detection tasks, the model needs to recover nonlinear 3D geometric information from 2D images, which poses high requirements for the multi-scale context awareness ability and key structure retention ability of features. However, standard SNN encoders (such as direct ANN2SNN conversion) lose a large amount of information during the spiking process. Therefore, in order to enhance the feature expression ability and reduce information loss, we propose a cross-scale fusion of attention coding unit named as CSGC,  as shown in Figure \ref{fig1}.

For CSGC, we design a parallel architecture including channel attention and spatial attention and a gate control unit. The Channel Attention addresses the issue of which feature channels are more important for 3D geometric reasoning. The Spatial Attention is used to focus on which areas in an image are more critical to the current target scale. The original input tensor $ x \in {R^{ C \times H \times W}}$ is firstly repeated at each time step to obtain information in the time dimension. Then, the original tensor is fed into both the channel attention and spatial attention modules. In the channel attention part, we use a Linear-ReLU-Linear structure to learn and update the weights in the channel dimension by the Equation (\ref{ca}):
\begin{equation}\label{ca}CA(x)=Linear\left(ReLU(Linear(x))\right).\end{equation}
The number of neurons in the middle layer of the network is set to the number of input channels /ratio, and the ratio is set to 4. This structure can effectively highlight important channel features, thereby enhancing the model's ability to perceive key information. In the spatial attention part, we conduct feature extraction on the same feature map by different convolution kernel (Equation (\ref{eqsa})). Small convolution kernels are employed to capture local subtle features for small objects, enabling the detection of small-sized objects. Conversely, larger convolution kernels are utilized to cover larger areas for detecting large objects and effectively capturing global information. To better integrate information from different scales, we employ three different sizes of convolution kernels and assign dynamic weights to the feature maps after different convolutional processing by the learnable parameters $\alpha$, $\beta$ and $\gamma$, respectively, with the initial weights all set to 0.5. At the same time, we invoke the residual idea of \cite{10} to connect the original feature map with the output part. This not only enhances the features but also prevents information loss, ensuring that the model can fully utilize all useful information from the original input. 
\begin{equation}\label{eqsa}\begin{aligned}
SA(x) = & Conv_{3\times3}\left(ReLU\left(Conv_{3\times3}(x)\right)\right)\cdot\alpha +\\
        &  Conv_{5\times5}\left(ReLU\left(Conv_{5\times5}(x)\right)\right)\cdot\beta +\\
        &  Conv_{7\times7}\left(ReLU\left(Conv_{7\times7}(x)\right)\right)\cdot\gamma +\\
        &  x.
\end{aligned}\end{equation}
\

The output spike information after the above channel and spatial attention processing is controlled and adjusted by a Sigmoid gating unit, which is defined as Equation (\ref{sig}):
\begin{equation}\label{sig}O(x)=\sigma\left(SA(x)\odot CA(x)\right),\end{equation} 
where $\sigma$ denotes the Sigmoid function, which maps values to the range (0,1), and $\odot$ denotes the Hadamard product. Then, it  performs the Hadamard product operation with the output of the spiking neuron, imitating the filtering function of the synapses, as defined by Equation (\ref{fi}):
\begin{equation}\label{fi}\widehat{O}(x)=O(x)\odot SNN(x),\end{equation}

where $O(x)$ represents the attention scores obtained after the parallel channel attention and spatial attention modules, indicating the importance of the detected features. $SNN(x)$ denotes the output spikes from the original feature x after processing through the Leaky Integrate-and-Fire (LIF) neuron.  Since in SNNS, features exist in the form of binary pulses, direct weighting would undermine their sparsity and event-driven characteristics. Therefore, CSGC takes the attention score $O(x)$  as the gate signal and performs a Hadamard product with the binary output (0 or 1) of $SNN(x)$, which is generated by an LIF neuron, allowing only the spikes in the high-attention region to pass through, thereby obtaining the final output $\widehat{O}(x)$. This essentially simulates the filtering mechanism in biological synapses, achieving feature screening while maintaining the sparsity of spikes.

\subsection{Light-weight Residual}
\begin{figure}[htbp]
    \centering
    \includegraphics[width=0.9\columnwidth]{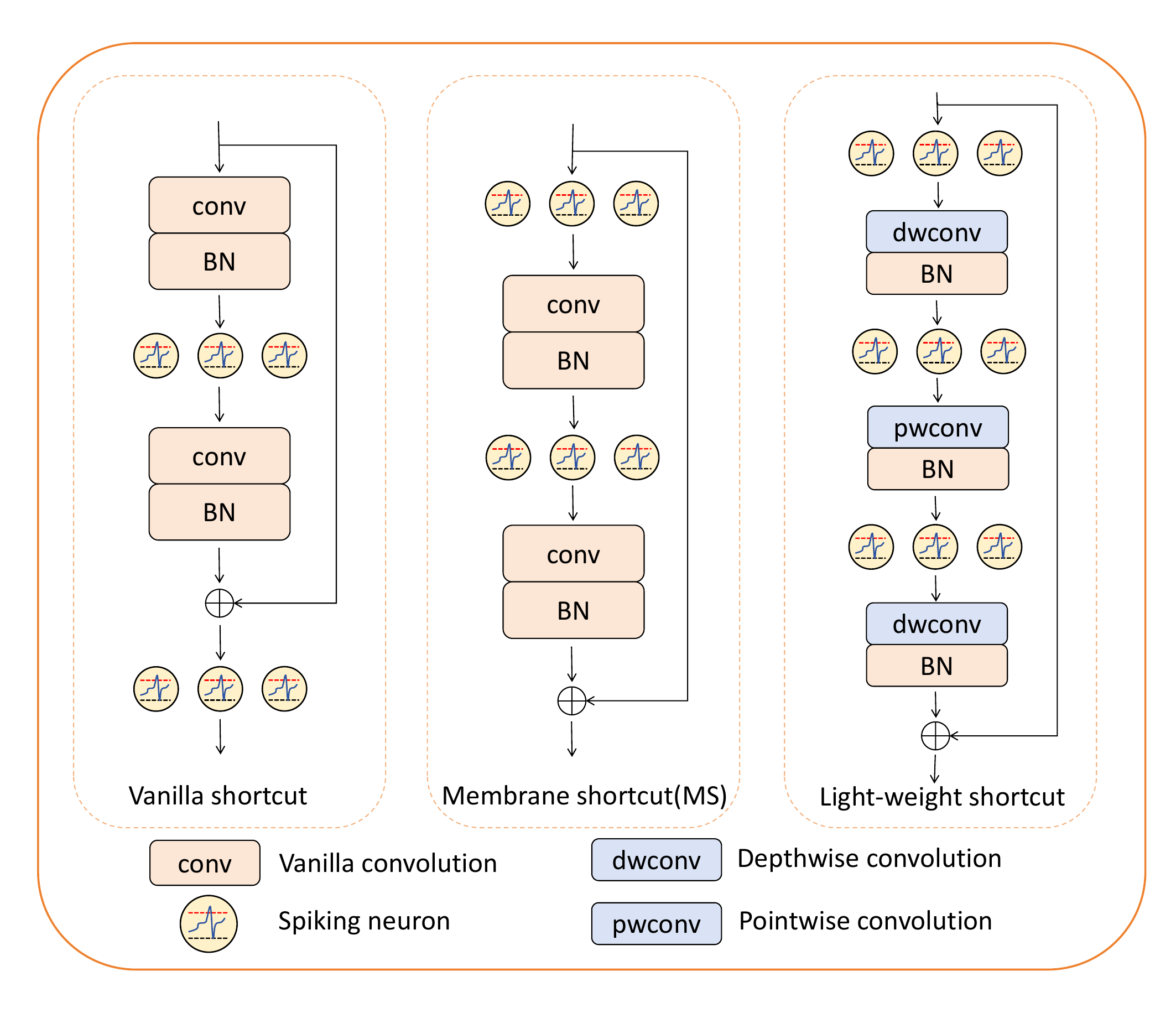} 
    \captionsetup{justification=raggedright, singlelinecheck=false} 
    \caption{\small \small Light-weight residual blocks. The rightmost figure uses depth-wise convolution and point-wise convolution to replace vanilla convolution. }
    \label{fig3}
\end{figure}
When SNN is applied to high-complexity tasks such as monocular 3D object detection, the residuals of standard convolution need to be repeatedly executed over multiple time steps, and the fully connected channel interaction  can lead  to a large amount of invalid computations, which goes against the original intention of energy efficiency of the SNN. Therefore, we introduce a novel light-weight residual block, as shown in Figure \ref{fig3}, is introduced to significantly reduce the number of model parameters and computational complexity by integrating depth-wise separable convolution technology and a membrane-based shortcut method, thereby enhancing the efficiency of model training. In terms of implementation details, the lightweight residual block first employs depth-wise convolution to perform independent convolution operations on each channel of the input data. Compared to traditional convolution, depth-wise convolution has the advantage of operating on each channel individually, thus avoiding redundant computations across channels and significantly improving computational efficiency. After completing the depth-wise convolution, we then apply point-wise convolution , with the purpose of integrating information from various channels and enhancing the expressive power of features. Then, we apply depth-wise convolution again to further extract and strengthen features. During this process, to maintain the flow of information and prevent the vanishing gradient problem, we introduce a shortcut path based on the membrane method

For regular membrane-based residual block, the computational volume $\alpha_1$ and number of parameters $\beta_1$ can be calculated by  expression (\ref{ms-co}) and  the expression (\ref{ms-pa}) respectively:
\begin{equation}
\label{ms-co}
\alpha_1=k\cdot k\cdot C_{in}\cdot C_{out}\cdot W_{out}\cdot H_{out}, 
\end{equation}
\begin{equation}
\label{ms-pa}
\beta_1=k\cdot k\cdot C_{in}\cdot C_{out}.
\end{equation}

When the above light-weight residual block is introduced, the computational volume $\alpha_2$ and parametric quantities $\beta_2$ can be calculated by the expression (\ref{af-co}) and the expression (\ref{af-pa}) respectively:
\begin{equation}
\label{af-co}
\alpha_2=k\cdot k\cdot C_{in}\cdot W_{out}\cdot H_{out}+C_{in}\cdot C_{out}\cdot W_{out}\cdot H_{out},
\end{equation}
\begin{equation}
\label{af-pa}
\beta_2=k\cdot k\cdot C_{in}+C_{in}\cdot C_{out}.
\end{equation}

For the SpikeSMOKE with light-weight residual block, the convolution kernel size $k$ is $3\times3$, $C_{\text{out}}=2C_{\text{in}}$, $W_{\text{out}}=\frac{1}{2}W_{\text{in}}$, $H_{\text{out}}=\frac{1}{2}H_{\text{in}}$. Then, the ratio of the computational effort can be calculated by the Equation (\ref{ratio-co}): 
\begin{equation}\label{ratio-co}\frac{C_{in}\cdot W_{out}\cdot H_{out}(3k^2+2C_{in})}{3k^2\cdot C_{in}\cdot C_{out}\cdot W_{out}\cdot H_{out}}=\frac{1}{2C_{in}}+\frac{1}{27}.\end{equation}
Similarly, the ratio of their parametric quantities is the Equation (\ref{ratio-pa}):
\begin{equation}\label{ratio-pa}\frac{k^2\cdot C_{in}+C_{in}\cdot C_{out}+k^2C_{out}}{k^2C_{in}\cdot C_{out}+k^2C_{out}\cdot C_{out}}=\frac{1}{2C_{in}}+\frac{1}{27}.\end{equation}

Observably, compared to the pre-improvement phase, the proposed light-weight residual block diminishes computational and parametric quantities by approximately 27 times when $C_{in}$ is large enough.  The exact reduction ratio depends on the $C_{in}$. This novel lightweight residual block, through carefully designed convolution operations and shortcut connections, achieves a reduction in computational load and parameter count while preserving model performance as much as possible, offering new perspectives for the optimization of deep learning models.

\section{Experiments}
\label{sec:experiments}

\subsection{Datasets}

\textbf{KITTI. }

The KITTI dataset is widely used in the field of autonomous driving and computer vision. The monocular 3D object detection utilizes single-camera data for stereo object detection across the Car, Pedestrian, and Cyclist categories, with 3,712 pictures in the training set and 3,769 pictures in the validation set, encompassing diverse scenarios such as city roads and highways. The detected objects include 3D bounding boxes, bird's eye view bounding boxes, 2D bounding boxes, and can be classified into Easy, Moderate, and Hard categories based on detection complexity. We used an evaluation metric called 11-point Interpolated Average Precision metric \cite{45}, which approximates the shape of the Precision/Recall curve, at each difficulty level. It is defined as the Equation (\ref{AP}):

\begin{equation}\label{AP}AP|N=\frac{1}{|N|}\sum_{n\in N}f_{inter}(n) 
\end{equation}

Where N denotes the set of recall levels that represent the exact intervals, this experiment takes $R_{11}=\{\frac{1}{10}, \frac{2}{10}, \frac{3}{10}, \cdots, 1\}$. $f_{inter}(n)$ denotes the precision obtained after interpolation calculation, instead of averaging the precision values, the maximum precision greater than or equal to the current recall is taken in calculating the precision.  It is defined as the Equation (\ref{Finter}):

\begin{equation}f_{inter}\label{Finter}(n)=\max_{n^{\prime}:n^{\prime}\geq n}f(n^{\prime})
\end{equation}

\renewcommand{\arraystretch}{1.45}
\begin{table*}[htbp]
\centering
\caption{\small The results of 3D object detection and detection under BEV are performed on\\ the validation set of KITTI dataset for the car class.} 
\label{tab1}
\setlength{\tabcolsep}{7pt}
\begin{tabular}{ccccccccccc}
\hline
\multicolumn{2}{c}{\multirow{2}{*}{Methods(IoU)}} & \multirow{2}{*}{Parameters(M)} & \multirow{2}{*}{Power(pJ)} & \multicolumn{3}{c}{3D Object Detection} & \multicolumn{3}{c}{Birds’ Eye View} & \multirow{2}{*}{Runtime(s)} \\ \cline{5-10}
\multicolumn{2}{c}{} &  &  & Easy & Moderate & Hard & Easy & Moderate & Hard &  \\ \hline
\multirow{7}{*}{ANN} & SMOKE(0.7) & 19.51 & 2.17E+11 & 14.76 & 12.85 & 11.5 & 19.99 & 15.61 & 15.28 & 0.03 \\
 & SMOKE*(0.5) & 19.51 & 2.17E+11 & 29.16 & 25.22 & 21.47 & 32.52 & 29.59 & 25.86 & 0.03 \\
 & SMOKE*(0.7) & 19.51 & 2.17E+11 & 12.03 & 11.14 & 10.92 & 17.97 & 13.08 & 12.06 & 0.03 \\
 & FCOS3D(0.7) & 50.62 & 3.04E+11 & 13.9 & 11.61 & 10.98 & 18.16 & 14.02 & 13.85 & -- \\
 & MonoDLE(0.7) & 20.82 & 2.87E+11 & 17.45 & 13.66 & 11.68 & 24.97 & 19.33 & 17.01 & 0.04 \\
 & MonoDTR(0.7) & 54.25 & 5.42E+11 & 24.52 & 18.57 & 15.51 & 33.33 & 25.35 & 21.68 & 0.037 \\
 & MonoLSS(0.7) & 21.51 & 2.81E+11 & 25.91 & 18.29 & 15.94 & 34.70 & 25.36 & 21.84 & 0.035 \\ \hline
\multirow{4}{*}{SNN} & SpikeSMOKE(0.7) & 19.51 & 5.97E+10 & 8.96 & 7.49 & 7.31 & 12.07 & 9.95 & 9.32 & 0.09 \\
 & \textbf{SpikeSMOKE-CSGC(0.7)} & \textbf{19.56} & \textbf{6.04E+10} & \textbf{11.78} & \textbf{10.69} & \textbf{10.48} & \textbf{15.67} & \textbf{13.68} & \textbf{11.83} & 0.105 \\
 & SpikeSMOKE(0.5) & 19.51 & 5.97E+10 & 20.87 & 19.72 & 16.89 & 27.64 & 22.47 & 21.65 & 0.09 \\
 & \textbf{SpikeSMOKE-CSGC(0.5)} & \textbf{19.56} & \textbf{6.04E+10} & \textbf{28.83} & \textbf{22.75} & \textbf{19.44} & \textbf{36.23} & \textbf{26.88} & \textbf{25.75} & 0.105 \\ \hline
\end{tabular}%
\end{table*}

\begin{table*}[htbp]
\centering
\caption{\small \small The results of 3D object detection and detection under BEV are performed on \\ the test set of KITTI dataset for the car class. }
\label{tab2}
\begin{tabular}{ccccccccc}
\toprule
{Methods} & {\begin{tabular}[c]{@{}c@{}}Parameters\\ (M)\end{tabular}} & {\begin{tabular}[c]{@{}c@{}}Power\\ (pJ)\end{tabular}} & \multicolumn{3}{c}{3D Object Detection} & \multicolumn{3}{c}{Birds’ Eye View} \\ \cline{4-9} 
 &  &  & Easy & Moderate & Hard & Easy & Moderate & Hard \\ \midrule
SMOKE-ANN* & 19.51 & 2.17E+11 & 28.97 & 23.57 & 20.21 & 30.48 & 27.76 & 25.81 \\
\textbf{SpikeSMOKE} & \textbf{19.51} & \textbf{5.97E+10} & \textbf{21.80} & \textbf{15.23} & \textbf{14.99} & \textbf{25.99} & \textbf{20.69} & \textbf{17.92} \\
SpikeSMOKE-L & 6.32 & 2.24E+10 & 19.46 & 10.52 & 10.23 & 24.01 & 15.74 & 15.37 \\
\begin{tabular}[c]{@{}c@{}}SpikeSMOKE -LCSGC\end{tabular} & 6.37 & 2.29E+10 & 20.64 & 15.32 & 12.93 & 25.15 & 18.98 & 18.08 \\ \bottomrule
\multicolumn{8}{l}{\parbox{0.66\linewidth}{\small The table is evaluated using the $AP|_{R_{11}}$ with a 0.5 IoU threshold.}} \\
\end{tabular}
\end{table*}

\begin{table}[htbp]
\centering
\caption{\small \small The results of 2D object detection is performed  on \\ the validation set of KITTI dataset for the car class. }
\label{tab3}
\setlength{\tabcolsep}{12pt}
\begin{tabular}{ccll}
\toprule
{Methods} & \multicolumn{3}{c}{2D Object Detection} \\ \cline{2-4} 
 & Easy & \multicolumn{1}{c}{Moderate} & \multicolumn{1}{c}{Hard} \\ \midrule
SMOKE-ANN* & 80.49 & 72.01 & 68.76 \\
SpikeSMOKE & 73.32 & 62.85 & 55.67 \\
\textbf{SpikeSMOKE-CSGC} & \textbf{75.58} & \textbf{65.49} & \textbf{64.37} \\
SpikeSMOKE-L & 51.59 & 44.01 & 42.71 \\
SpikeSMOKE-LCSGC & 52.33 & 49.04 & 43.04 \\ \bottomrule
\end{tabular}
\end{table}

\begin{table}[htbp]
\centering
\caption{\small \small Classification results on the classification task \\ dataset CIFAR-10. }
\label{tab4}
\setlength{\tabcolsep}{7pt}
\begin{tabular}{cccc}
\toprule
Architecture & \begin{tabular}[c]
{@{}c@{}}Coding\\ Schemes\end{tabular} & \begin{tabular}[c]{@{}c@{}}Time\\ Steps\end{tabular} & \begin{tabular}[c]{@{}c@{}}CIFAR10\\ Acc. (\%)\end{tabular} \\ \midrule
ResNet-19{}\cite{8}{} & Phase Coding & 8 & 91.40 \\
VGG-16{}\cite{8}{} & Temporal Coding & 100 & 92.68 \\
ResNet-19{}\cite{8}{} & Rate Coding & 6 & 93.16 \\
MS-ResNet-18 & Direct Coding & 6 & 94.92 \\
\textbf{MS-ResNet-18} & \textbf{CSGC Coding} & \textbf{6} & \textbf{95.98(+1.06)} \\ \bottomrule
\end{tabular}
\end{table}

\begin{table}[!h]
\centering
\caption{\small \small Classification results on the classification task \\dataset CIFAR-100. }
\label{tab5}
\setlength{\tabcolsep}{2pt}
\begin{tabular}{cccccc}
\toprule
Methods & Architecture & \begin{tabular}[c]
{@{}c@{}}Spike\end{tabular} & \begin{tabular}[c]{@{}c@{}}Params\\ (M)\end{tabular} & \begin{tabular}[c]{@{}c@{}}Time\\ Steps\end{tabular} & \begin{tabular}[c]{@{}c@{}}CIFAR100\\ Acc. (\%)\end{tabular} \\ \midrule
ANN{}\cite{8}{} & MS-ResNet-18 & $\times$ & 12.54 & N/A & 80.67 \\ \midrule
MS-ResNet-SNN{}\cite{8}{} & MS-ResNet-18 & $\checkmark$ & 12.54 & 6 & 76.41 \\ \midrule
\multirow{2.2}{5.2em}{CSGC-SNN}
 & MS-ResNet-18 & $\checkmark$ & 12.72 & 6 & 79.58 \\
 & MS-ResNet-18 & $\checkmark$ & 12.72 & 4 & 77.97 \\ \bottomrule

\end{tabular}
\end{table}

\textbf{CIFAR-10/100. }

The CIFAR-10 and CIFAR-100 datasets are commonly used for image classification. The CIFAR-10 dataset contains 10 categories of color images, each containing 6000 images of size 32x32 pixels, for a total of 60, 000 images. The CIFAR-100 dataset extends CIFAR-10 with 100 categories, each containing 600 images of size 32x32 pixels. contains 600 images, for a total of 60, 000 images with image size of 32x32 pixels.

\textbf{NuScenes-mini. }

The NuScenes-mini dataset is a subset of the NuScenes dataset, which is collected by 6 cameras, 1 Lidar, 5 millimetre wave radars, GPS and IMU, covering complex traffic conditions as well as variations in weather and light. The dataset contains 8 attributes of 3 types of objects, namely vehicles, bicycles, and pedestrians, and 10 scenes with continuous images of about 20 seconds each, which is an important dataset for 3D object detection in the field of autonomous driving. The dataset presents a composite metric, the nuScenes Detection Score (NDS), which was combined by calculating weighted sums (mAP, mATE, mASE, mAOE, mAVE, and mAAE).

\subsection{Experiments Setup}
We used random horizontal flip, random scale and shift as data augmentation aiming to increase the diversity of training data. In the network design, we set the number of groups for group normalization to 32 and for less than 32, we take 16. based on the analysis in literature \cite{44}, we set $\begin{bmatrix}\bar{h, }\bar{w, }\overline{l, }
\end{bmatrix}^\top=[1.63, 1.53, 3.88]^\top and \begin{bmatrix}
\mu_z , \sigma_z
\end{bmatrix}^\top=[28.01{\operatorname*{, 16.32}}]^\top. $ The input image resolution size of the network is 1280x384, and after multiple downsampling, the size is reduced to 32 times of the original size. We trained 172 epoch  using four NVIDIA GeForce RTX 4090 GPU, with a size of 4 per batch and a learning rate set to $1.25\times10^{-4}$. The decay strategy for the learning rate was to reduce the learning rate to 10 times of the original at epoch 47 and 90. We used a threshold of 0.25 to filter the detection object. The lightweight strategy adopted in the experiment only acts on the backbone of feature extraction. Additionally, to more clearly present the configuration of each layer in the network, we have used a tabular format for detailed explanation, as shown in Table \ref{tab8}. The table lists key information including the size of the convolutional kernels, the number of input and output channels, the type of neurons, and the dimensions of the input and output, all of which are essential parameters when constructing a deep learning model.

\begin{figure}[htbp]
    \centering
\includegraphics[width=0.9\columnwidth]{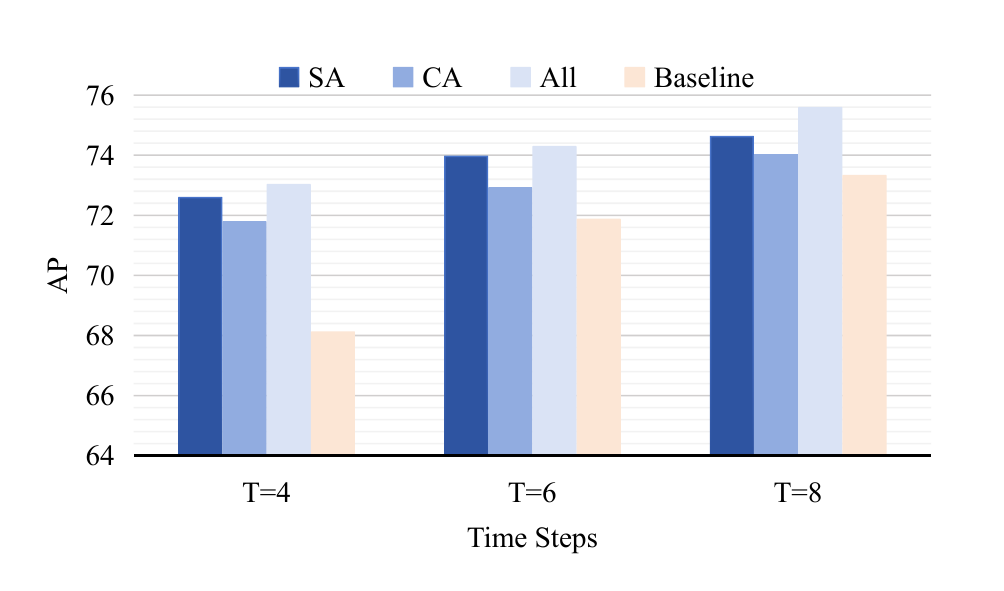} 
    \captionsetup{justification=raggedright, singlelinecheck=false} 
    \caption{\small \small Ablation experiments in CSGC with different attention modules at different time steps. }
    \label{fig5}
\end{figure}

\begin{table}[]
\centering
\caption{Results on the validation set of the NuScenes-mini dataset.} 
\label{tab6}
\begin{tabular}{>{\centering\arraybackslash}m{2.5cm} >{\centering\arraybackslash}m{1.5cm} >{\centering\arraybackslash}m{1.5cm}}
\toprule 
Methods & mAP & NDS \\
\midrule 
SMOKE & 31.3 & 37.2 \\
SpikeSMOKE & 25.3 & 29.8 \\
SpikeSMOKE-CSGC & 26.6 & 31.2 \\
\bottomrule 
\end{tabular}
\end{table}

\begin{figure*}[htbp]
    \centering
    \makebox[\textwidth][c]{\includegraphics[width=1\textwidth]{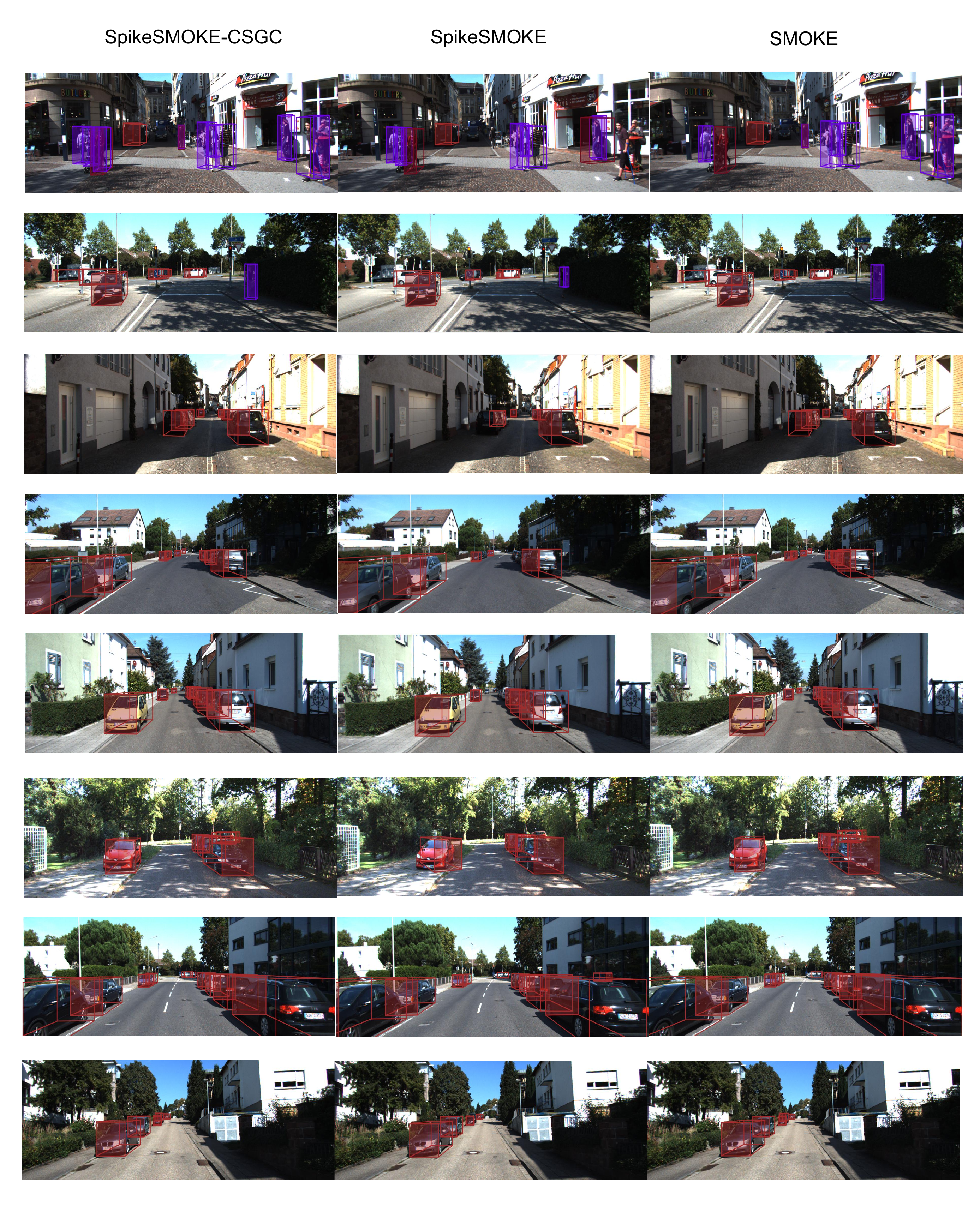}}%

    \caption{\small \small The visualization results from monocular 3D object detection on the KITTI dataset present a clear and intuitive representation of the performance and accuracy of the detection algorithm in identifying and localizing 3D objects within the complex road scenarios depicted in the dataset. }
    \label{fig4}
\end{figure*}

\begin{table*}[!h]
\centering
\caption{\small \small Ablation experiments on the threshold of spike neurons. }
\label{tab7}
\begin{tabular}{cccccccc}
\toprule
{Methods} & {vth=0.25} & {vth=0.5} & {vth=0.75} & {vth=1.0} & \multicolumn{3}{c}{2D Object Detection} \\ \cline{6-8} 
 &  &  &  &  & Easy & Moderate & Hard \\ \midrule \multirow{4}{9.5em}{SpikeSMOKE-CSGC}
 & $\checkmark$ &  &  &  & 64.31 & 53.84 & 51.62 \\
 &  & $\checkmark$ &  &  & 72.06 & 61.27 & 60.67 \\
 &  &  & $\checkmark$ &  & \textbf{75.58} & \textbf{65.49} & \textbf{64.37} \\
 &  &  &  & $\checkmark$ & 54.10 & 45.88 & 44.98 \\ \bottomrule
\end{tabular}
\end{table*}

\begin{table*}[!h]
\renewcommand{\arraystretch}{1.45}
\centering
\caption{\small The detailed settings of the SpikeSMOKE-CSGC architecture.}
\label{tab8}
\begin{tabular}{ccccc}
\hline
\multicolumn{2}{c}{\textbf{Layer Name}} & \textbf{Input Size} & \textbf{Layer} & \textbf{Output Size} \\ \hline
 & Base\_layer & (T,B,3,384,1280) & 7×7Conv,32,stride1,padding3→GN→LIF & (T,B,32,384,1280) \\
 & CSGC & (T,B,32,384,1280) & Channel-Attn + Spatial-Attn (3×3/5×5/7×7Conv) + Sigmoid & (T,B,32,384,1280) \\
 & Avgpooling & (T,B,32,384,1280) & AvgPool2d(kernel=2, stride=2) & (T,B,32,192,640) \\
 & Layer0 & (T,B,32,192,640) & 3×(LIF→3×3Conv→LIF→3×3Conv)-LIF & (T,B,64,96,320) \\
 & Layer1 & (T,B,64,96,320) & 4×(LIF→3×3Conv→LIF→3×3Conv)-LIF & (T,B,128,48,160) \\
 & Layer2 & (T,B,128,48,160) & 6×(LIF→3×3Conv→LIF→3×3Conv)-LIF & (T,B,256,24,80) \\
\multirow{-7}{*}{\textbf{Backbone}} & Layer3 & (T,B,256,24,80) & 4×(LIF→3×3Conv→LIF→3×3Conv)-LIF & (T,B,512,12,40) \\ \hline
 & Upblock0 & -- & LIF→3×3Conv(512-256)→LIF→ConvTranspose(2×) & -- \\
 & Upblock1 & -- & LIF→3×3Conv(256-128)→LIF→ConvTranspose(2×) & -- \\
\multirow{-3}{*}{\textbf{Neck}} & Upblock2 & -- & LIF→3×3Conv(128-64)→LIF→ConvTranspose(2×) & (T,B,64,96,320) \\ \hline
 & class head & (T,B,64,96,320) & LIF→Conv3×3(64→256)→GN→LIF→Conv1×1(256→3) & (B,3,96,320) \\
\multirow{-2}{*}{\textbf{Head}} & regression head & (T,B,64,96,320) & LIF→Conv3×3(64→256)→GN→LIF→Conv1×1(256→8) & (B,8,96,320) \\ \hline
\end{tabular}
\end{table*}

\subsection{Main Results}

\textbf{Object Detection Performance on KITTI. }

We validate the performance on the validation and test sets of the KITTI dataset for 3D object detection and bird's-eye view under three levels of Easy, Moderate and Hard.  * denotes the experimental results obtained by us after configuring the environment and training based on the SMOKE. The 3d object detection of our proposed SpikeSMOKE with CSGC (SpikeSMOKE-CSGC) can reach 28.83/11.78 (Easy), 22.75/10.69 (Moderate), 19.44/10.48 (Hard), and for the BEV detection, they can reach  36.23/15.67 (Easy), 26.88/13.68 (Moderate), 25.75/11.83 (Hard), as shown in Table \ref{tab1}, where the metrics is evaluated by $AP|_{R_{11}}$ at 0.5/0.7 IoU threshold. From the results of these experiments, it can be found that although our detection performance has a little gap compared with the SMOKE-ANN*, energy consumption has been significantly reduced.  For example, we calculate the energy consumption of the hard category for the 0.7 IoU threshold and find that it can be reduced by 72.2\%, while the detection performance is reduced by only 4\%. These experimental results are obtained because SMOKE-ANN* relies on continuous value activation and can precisely model 3D geometric attributes. However, SpikeSMOKE converts features into binary spike sequences (0/1), and information is inevitably lost during the encoding process. Meanwhile, the low-power consumption feature of our SNN event-driven mechanism reduces energy efficiency. The formula for calculating energy consumption is as follows:
\begin{equation}\begin{aligned}
EC_{SNN} & =Synapsed_{activated}^{SNN}\times0.9, \\
EC_{ANN} & =FLOPS^{ANN}\times4.6,
\end{aligned}\end{equation}
where 0.9 denotes the consumption per accumulator operation and 4.6 denotes the multiplier word operation \cite{027}\cite{028}. With a time step T=8, we achieved a runtime latency of 0.105s when performing inference on 4 NVIDIA GeForce RTX 4090 GPUs. Since we extended the time dimension when training on the GPU, it is reasonable that the latency would be larger compared to the traditional ANNs model.

It is well known that monocular 3D object detection is commonly used in resource-constrained scenarios such as edge devices, embedded devices, etc., therefore, lightweighting is an important part of research.
Based on this, we further discuss SpikeSMOKE-CSGC's lightweighting by light-weight residul block. The experiments show that the number of parameters is only 1/3 of SMOKE-ANN*, and the amount of computation is only 1/10, as shown in Table \ref{tab2}. As a result, the SpikeSMOKE model may provide a new effective solution to reduce the energy consumption of monocular 3D object detection and improve its energy efficiency.

Comparing with the baseline SpikeSMOKE, the proposed SpikeSMOKE-CSGC shows significant improvements in both 3D object detection and BEV detection, with gains of 2.82, 3.2, and 3.17 for 3D object detection, and 3.6, 3.73, and 2.51 for BEV detection, as displayed in Table \ref{tab1}. The results of the SpikeSMOKE-CSGC with lightweight treatment (SpikeSMOKE-LCSGC) can also demonstrate notable enhancements  compared to that of the SpikeSMOKE-L, as indicated in Table \ref{tab2}. 
Therefore, the improved CSGC method  has a significant effect on 2D/3D object detection.

\textbf{Efficiency and Generalizability Validation for CSGC. }

We validate the classification results of the proposed CSGC coding strategy on MS-ResNet18 on the classification dataset CIFAR-10/100, as shown in Table \ref{tab4} and Table \ref{tab5}. We note that the accuracy rate of the MS-ResNet18 with CSGC coding on the CIFAR-10 is 1.06\% higher than the direct coding method. On the CIFAR-100, the proposed method MS-ResNet18-SNN with CSGC coding can achieve 79.58\% by 6 time steps, which is 3.17\% higher than MS-ResNet18-SNN. Meanwhile, we validated its effectiveness on the NuScenes-mini dataset, as shown in Table \ref{tab6}. Consequently, our proposed CSGC coding method is  effective and generalized.

\textbf{Ablation Studies on the Effects of Various Attentions. }

We performed ablation experiments on 2D object detection about spatial attention (SA) and channel attention (CA) of CSGC at time steps of 4, 6, and 8, respectively, as shown in Figure \ref{fig5}. We observe that the detection performance is better than that of the baseline when SA or CA is used, which means that each module of CSGC is effective. Obviously, when CA and SA are used together, their detection results are better. Additionally, as the time step increases, the detection performance gets better.

\textbf{Ablation Studies of Individual Neuronal Thresholds. }

Since the threshold of the spike neurons has a great impact on the detection performance of the model, we conduct some ablation experiments on 2D detection by different neuron thresholds in order to get the most suitable LIF neuron threshold. Based on the experiment results, we obtain that the best performance  is obtained when the neuron threshold  vth=0.75, which is used in this paper, as shown in Table \ref{tab7}.

\textbf{Qualitative Results. }

We provide visualization results of the monocular 3D object detection on the KITTI dataset and compare them with the SpikeSMOKE and SMOKE, as shown in Figure \ref{fig4}. It can be observed that the detection performance of the Spikesmoke-CSGC architecture is superior to that of the SpikeSMOKE architecture and is also closer to that of the SMOKE architecture. Through these results, we can visually understand the model's capability to accurately identify and locate 3D objects in complex road scenarios, thereby affirming the significant potential application of this technology in areas such as autonomous driving.

\section{Conclusion}

\label{sec:conclusion}
With the widespread use of 3D object detection in applications such as autonomous driving, the low energy consumption problem is getting more and more attention. As widely acknowledged, low-power consumption stands out as a key feature of brain-like SNNs, offering a potential new solution for energy-efficient 3D object detection. Based on the SMOKE architecture and SNNs model, we have constructed a novel Spiking SMOKE architecture named SpikeSMOKE for monocular 3D object detection to reduce its energy consumption. In this architecture,  the discrete signaling characteristics of SNNs may result in information loss and restrict their capacity for feature representation. Therefore,  we have proposed a new cross-scale gating coding module named CSGC that can enhance the feature expression ability of the model, inspired by the synaptic filtering process in biological neurons. Furthermore, we have also proposed  a lightweight residual block to reduce the computational effort while maintaining the impulse computation paradigm. We have conducted a large number of experiments on the KITTI, NuScenes-mini and CIFAR10/100 datasets to verify that the SpikeSMOKE architecture and CSGC module proposed in this paper are effective. The experimental results on the KITTI dataset, show that SpikeSMOKE can achieve higher energy efficiency compared to SMOKE, e.g., 72.2\% reduction in energy consumption on Hard category, while the detection performance drops by only 4\%. Moreover, the experimental results also show that the CSGC-based SpikeSMOKE can achieve significant improvement over the baseline SpikeSMOKE.  SpikeSMOKE-L (lightweight) can further reduce the amount of parameters by 3 times and computation by 10 times compared to SMOKE. In the CIFAR-10/100 classification task, the CSGC encoding strategy improves the correctness by 1.06\% and 3.17\%, respectively, validating its generality. 
Overall, the SpikeSMOKE architecture and CSGC mechanism can provide an efficient and feasible solution for low-power monocular 3D object detection. In the future, we will continue to explore how to utilize the SNNs model to further enhance the performance of monocular 3D object detection and further reduce its energy consumption.

\ifCLASSOPTIONcaptionsoff
  \newpage
\fi

\bibliographystyle{IEEEtran}
\bibliography{ref}

@InProceedings{16,
author={Krizhevsky, Alex and Sutskever, Ilya and Hinton, Geoffrey E},
booktitle={Advances in Neural Information Processing Systems (NeurIPS)},
pages={1--9},
title={Imagenet classification with deep convolutional neural networks},
volume={25},
year={2012},
month={Dec.},
}

@article{17,
  title={Deep Speech: Scaling up end-to-end speech recognition},
  author={Hannun, A},
  journal={arXiv preprint arXiv:1412.5567},
  year={2014}
}

@article{48,
  title={Spiking deep residual networks},
  author={Hu, Yangfan and Tang, Huajin and Pan, Gang},
  journal={IEEE Transactions on Neural Networks and Learning Systems},
  volume={34},
  number={8},
  pages={5200--5205},
  year={2021},
  publisher={IEEE}
}

@inproceedings{1,
 author = {Ren, Dayong and Ma, Zhe and Chen, Yuanpei and Peng, Weihang and Liu, Xiaode and Zhang, Yuhan and Guo, Yufei},
 booktitle = {Advances in Neural Information Processing Systems (NeurIPS)},
 
 pages = {41797--41808},

 title = {Spiking PointNet: Spiking Neural Networks for Point Clouds},
 volume = {36},
 year = {2023},
 month={Dec.},
}

@InProceedings{2,
author = {Liu, Zechen and Wu, Zizhang and Toth, Roland},
title = {SMOKE: Single-Stage Monocular 3D Object Detection via Keypoint Estimation},
booktitle = {Proceedings of the IEEE/CVF Conference on Computer Vision and Pattern Recognition (CVPR)},
pages={996--997},
month = {Jun.},
year = {2020}
}

@InProceedings{3,
author="Luo, Xinhao
and Yao, Man
and Chou, Yuhong
and Xu, Bo
and Li, Guoqi",

title="Integer-Valued Training and Spike-Driven Inference Spiking Neural Network for High-Performance and Energy-Efficient Object Detection",
booktitle="European Conference on Computer
Vision (ECCV)",
year="2024",
month={Sep.},

pages="253--272",

}

@inproceedings{7,
  title={Temporal-coded deep spiking neural network with easy training and robust performance},
  author={Zhou, Shibo and Li, Xiaohua and Chen, Ying and Chandrasekaran, Sanjeev T and Sanyal, Arindam},
  booktitle={Proceedings of the AAAI conference on artificial intelligence (AAAI)},
  volume={35},
  number={12},
  pages={11143--11151},
  year={2021},
  month={May.},
}

@inproceedings{8,
  title={Gated attention coding for training high-performance and efficient spiking neural networks},
  author={Qiu, Xuerui and Zhu, Rui-Jie and Chou, Yuhong and Wang, Zhaorui and Deng, Liang-jian and Li, Guoqi},
  booktitle={Proceedings of the AAAI Conference on Artificial Intelligence (AAAI)},
  volume={38},
  number={1},
  pages={601--610},
  year={2024},
month={Feb.}
}

@article{9,
  title={Mobilenets: Efficient convolutional neural networks for mobile vision applications},
  author={Howard, Andrew G},
  journal={arXiv preprint arXiv:1704.04861},
  year={2017}
}

@inproceedings{10,
  title={Deep residual learning for image recognition},
  author={He, Kaiming and Zhang, Xiangyu and Ren, Shaoqing and Sun, Jian},
  booktitle={Proceedings of the IEEE Conference on Computer Vision and Pattern
Recognition (CVPR)},
  pages={770--778},
  year={2016},
 month={Jun.}
}

@inproceedings{
11,
title={Optimal {ANN}-{SNN} Conversion for High-accuracy and Ultra-low-latency Spiking Neural Networks},
author={Tong Bu and Wei Fang and Jianhao Ding and PENGLIN DAI and Zhaofei Yu and Tiejun Huang},
booktitle={International Conference on Learning Representations (ICLR)},
year={2022},
month={Apr.},
pages = {1-19},
}

@InProceedings{12,
author="Oh, Youngmin
and Kim, Hyung-Il
and Kim, Seong Tae
and Kim, Jung Uk",

title="MonoWAD: Weather-Adaptive Diffusion Model for Robust Monocular 3D Object Detection",
booktitle="European Conference on Computer Vision (ECCV)",
year="2024",
month="Sep.",
pages="326--345",

}

@InProceedings{14,
author="Kim, Sanmin
and Kim, Youngseok
and Hwang, Sihwan
and Jeong, Hyeonjun
and Kum, Dongsuk",

title="LabelDistill: Label-Guided Cross-Modal Knowledge Distillation for Camera-Based 3D Object Detection",
booktitle="European Conference on Computer Vision (ECCV)",
year="2024",
month={Sep.},
pages="19--37"}

@InProceedings{15, 
title={Mono3DVG: 3D Visual Grounding in Monocular Images}, 
volume={38}, 

DOI={10.1609/aaai.v38i7.28525}, 
abstractNote={We introduce a novel task of 3D visual grounding in monocular RGB images using language descriptions with both appearance and geometry information. Specifically, we build a large-scale dataset, Mono3DRefer, which contains 3D object targets with their corresponding geometric text descriptions, generated by ChatGPT and refined manually. To foster this task, we propose Mono3DVG-TR, an end-to-end transformer-based network, which takes advantage of both the appearance and geometry information in text embeddings for multi-modal learning and 3D object localization. Depth predictor is designed to explicitly learn geometry features. The dual text-guided adapter is proposed to refine multiscale visual and geometry features of the referred object. Based on depth-text-visual stacking attention, the decoder fuses object-level geometric cues and visual appearance into a learnable query. Comprehensive benchmarks and some insightful analyses are provided for Mono3DVG. Extensive comparisons and ablation studies show that our method significantly outperforms all baselines. The dataset and code will be released.}, 
number={7}, 
booktitle={Proceedings of the AAAI Conference on Artificial Intelligence (AAAI)}, author={Zhan, Yang and Yuan, Yuan and Xiong, Zhitong}, year={2024}, month={Feb.}, pages={6988-6996} }

@inproceedings{18,
  title={Bert: Pre-training of deep bidirectional transformers for language understanding},
  author={Kenton, Jacob Devlin Ming-Wei Chang and Toutanova, Lee Kristina},
  booktitle={Proceedings of naacL-HLT},
  volume={1},
  pages={2-18},
  year={2019},
}

@InProceedings{21,
    author    = {Kumar, Abhinav and Brazil, Garrick and Liu, Xiaoming},
    title     = {GrooMeD-NMS: Grouped Mathematically Differentiable NMS for Monocular 3D Object Detection},
    booktitle = {Proceedings of the IEEE/CVF Conference on Computer Vision and Pattern Recognition (CVPR)},
    month     = {Jun.},
    year      = {2021},
    pages     = {8973-8983}
}

@inproceedings{23,
  title={Monodgp: Monocular 3D object detection with decoupled-query and geometry-error priors},
  author={Pu, Fanqi and Wang, Yifan and Deng, Jiru and Yang, Wenming},
  booktitle={Proceedings of the Computer Vision and Pattern Recognition Conference (CVPR)},
  pages={6520--6530},
  year={2025},
  month={Jun.}
}

@inproceedings{24,
  title={MonoCD: Monocular 3D Object Detection with Complementary Depths},
  author={Yan, Longfei and Yan, Pei and Xiong, Shengzhou and Xiang, Xuanyu and Tan, Yihua},
  booktitle={Proceedings of the IEEE/CVF Conference on Computer Vision and Pattern Recognition (CVPR)},
  pages={10248--10257},
  year={2024},
month={Jun.}
}

@INPROCEEDINGS{27,
  author={Li, Zhenjia and Jia, Jinrang and Shi, Yifeng},
  booktitle={2024 International Conference on 3D Vision (3DV)}, 
  title={MonoLSS: Learnable Sample Selection For Monocular 3D Detection}, 
  year={2024},
  month={Mar.},
  volume={},
  number={},
  pages={1125-1135},
}

@inproceedings{
28,
title={Sparse Spiking Neural Network: Exploiting Heterogeneity in Timescales for Pruning Recurrent {SNN}},
author={Biswadeep Chakraborty and Beomseok Kang and Harshit Kumar and Saibal Mukhopadhyay},
booktitle={International Conference on Learning Representations (ICLR)},
year={2024},
month={May.},
pages={1--32}

}

@article{29,
  title={Brain-Inspired Efficient Pruning: Exploiting Criticality in Spiking Neural Networks},
  author={Chen, Shuo and Liu, Boxiao and You, Haihang},
  journal={arXiv preprint arXiv:2311.16141},
  year={2023}
}

@inproceedings{30,
  title={Spiking-diffusion: Vector quantized discrete diffusion model with spiking neural networks},
  author={Liu, Mingxuan and Gan, Jie and Wen, Rui and Li, Tao and Chen, Yongli and Chen, Hong},
  booktitle={2024 5th International Conference on Computer, Big Data and Artificial Intelligence (ICCBD + AI)},
  pages={627--631},
  year={2024},
  month={Nov.},
}

@article{32,
  title={Self-Distillation Learning Based on Temporal-Spatial Consistency for Spiking Neural Networks},
  author={Zuo, Lin and Ding, Yongqi and Jing, Mengmeng and Yang, Kunshan and Yu, Yunqian},
  journal={arXiv preprint arXiv:2406.07862},
  year={2024}
}

@InProceedings{33,
author = {Chollet, Francois},
title = {Xception: Deep Learning With Depthwise Separable Convolutions},
booktitle = {Proceedings of the IEEE Conference on Computer Vision and Pattern Recognition (CVPR)},
pages={1251--1258},
month = {Jul.},
year = {2017}
}

@inproceedings{34,
  title={Encoder-decoder with atrous separable convolution for semantic image segmentation},
  author={Chen, Liang-Chieh and Zhu, Yukun and Papandreou, George and Schroff, Florian and Adam, Hartwig},
  booktitle={Proceedings of the European conference on computer vision (ECCV)},
  pages={801--818},
  year={2018},
month={Sep.}
}

@article{38,
      title={Deep neural networks with weighted spikes},
  author={Kim, Jaehyun and Kim, Heesu and Huh, Subin and Lee, Jinho and Choi, Kiyoung},
  journal={Neurocomputing},
  volume={311},
  pages={373--386},
  year={2018},
  publisher={Elsevier}
}

@article{39,
  title={Rate coding versus temporal order coding: what the retinal ganglion cells tell the visual cortex},
  author={Van Rullen, Rufin and Thorpe, Simon J},
  journal={Neural computation},
  volume={13},
  number={6},
  pages={1255--1283},
  year={2001},
}

@INPROCEEDINGS{40,
  author={Windhager, Daniel and Moser, Bernhard A. and Lunglmayr, Michael},
  booktitle={2024 IEEE 6th International Conference on AI Circuits and Systems (AICAS)}, 
  title={SNN Architecture for Differential Time Encoding Using Decoupled Processing Time}, 
  year={2024},
  month={Apr.},
  volume={},
  number={},
  pages={26-30},
}

@inproceedings{43,
  title={Centernet: Keypoint triplets for object detection},
  author={Duan, Kaiwen and Bai, Song and Xie, Lingxi and Qi, Honggang and Huang, Qingming and Tian, Qi},
  booktitle={Proceedings of the IEEE/CVF International Conference on Computer Vision (ICCV)},
  pages={6569--6578},
  year={2019},
month = {Oct.},
}

@inproceedings{44,
  title={Disentangling monocular 3d object detection},
  author={Simonelli, Andrea and Bulo, Samuel Rota and Porzi, Lorenzo and L{\'o}pez-Antequera, Manuel and Kontschieder, Peter},
  booktitle={Proceedings of the IEEE/CVF International Conference on Computer Vision (ICCV)},
  pages={1991--1999},
  year={2019},
  month={Oct.}
}

@InProceedings{45,
author="Robertson, Stephen",
title="On Smoothing Average Precision",
booktitle="European Conference on Information Retrieval (ECIR)",
year="2012",
month={Dec.},
pages="158--169",
}

@ARTICLE{46,
  author={Hu, Yifan and Deng, Lei and Wu, Yujie and Yao, Man and Li, Guoqi},
  journal={IEEE Transactions on Neural Networks and Learning Systems}, 
  title={Advancing Spiking Neural Networks Toward Deep Residual Learning}, 
  year={2024},
  volume={32},
  number={2},
  pages={1-15},
  }

@ARTICLE{01,
  author={Yao, Man and Zhao, Guangshe and Zhang, Hengyu and Hu, Yifan and Deng, Lei and Tian, Yonghong and Xu, Bo and Li, Guoqi},
  journal={IEEE Transactions on Pattern Analysis and Machine Intelligence}, 
  title={Attention Spiking Neural Networks}, 
  year={2023},
  volume={45},
  number={8},
  pages={9393-9410},
  }

@article{03,
author = {Wei Fang  and Yanqi Chen  and Jianhao Ding  and Zhaofei Yu  and Timothée Masquelier  and Ding Chen  and Liwei Huang  and Huihui Zhou  and Guoqi Li  and Yonghong Tian },
title = {SpikingJelly: An open-source machine learning infrastructure platform for spike-based intelligence},
journal = {Science Advances},
volume = {9},
number = {40},
pages = {eadi1480-1-18},
year = {2023}}

@article{04,
  title={Temporal dendritic heterogeneity incorporated with spiking neural networks for learning multi-timescale dynamics},
  author={Zheng, Hanle and Zheng, Zhong and Hu, Rui and Xiao, Bo and Wu, Yujie and Yu, Fangwen and Liu, Xue and Li, Guoqi and Deng, Lei},
  journal={Nature Communications},
  volume={15},
  number={1},
  pages={277-297},
  year={2024},
}

@inproceedings{06,
 author = {Fang, Wei and Yu, Zhaofei and Chen, Yanqi and Huang, Tiejun and Masquelier, Timoth\'{e}e and Tian, Yonghong},
 booktitle = {Advances in Neural Information Processing Systems (NeurIPS)},
 pages = {21056--21069},
 title = {Deep Residual Learning in Spiking Neural Networks},
 volume = {34},
 year = {2021},
month={Dec.}
}

@inproceedings{07,
 author = {Zhu, Yaoyu and Yu, Zhaofei and Fang, Wei and Xie, Xiaodong and Huang, Tiejun and Masquelier, Timoth\'{e}e},
 booktitle = {Advances in Neural Information Processing Systems (NeurIPS)},
 pages = {30528--30541},
 title = {Training Spiking Neural Networks with Event-driven Backpropagation},
 volume = {35},
 year = {2022},
month={Dec.}
}

@inproceedings{08,
  title={Going deeper with directly-trained larger spiking neural networks},
  author={Zheng, Hanle and Wu, Yujie and Deng, Lei and Hu, Yifan and Li, Guoqi},
  booktitle={Proceedings of the AAAI conference on artificial intelligence (AAAI)},
  pages={11062--11070},
month={Feb.},
  year={2021},

}

@inproceedings{09,
title={Spikformer: When Spiking Neural Network Meets Transformer },
author={Zhaokun Zhou and Yuesheng Zhu and Chao He and Yaowei Wang and Shuicheng YAN and Yonghong Tian and Li Yuan},
booktitle={The Eleventh International Conference on Learning Representations (ICLR) },
year={2023},
month={May.}
}

@inproceedings{010,
 author = {Yao, Man and Hu, JiaKui and Zhou, Zhaokun and Yuan, Li and Tian, Yonghong and Xu, Bo and Li, Guoqi},
 booktitle = {Advances in Neural Information Processing Systems(NeruIPS)},
 pages = {64043--64058},
 title = {Spike-driven Transformer},
 volume = {36},
 year = {2023},
month={Dec.}
}

@inproceedings{011,
title={Spike-driven Transformer V2: Meta Spiking Neural Network Architecture Inspiring the Design of Next-generation Neuromorphic Chips},
author={Man Yao and JiaKui Hu and Tianxiang Hu and Yifan Xu and Zhaokun Zhou and Yonghong Tian and Bo XU and Guoqi Li},
booktitle={The Twelfth International Conference on Learning Representations (ICLR)},
year={2024},
month={May.}
}

@InProceedings{012,
author="Lin, Hongbin
and Zhang, Yifan
and Niu, Shuaicheng
and Cui, Shuguang
and Li, Zhen",
title="MonoTTA: Fully Test-Time Adaptation for Monocular 3D Object Detection",
booktitle="European Conference on Computer
Vision (ECCV)",
year="2025",
pages="96--114",
isbn="978-3-031-72784-9"
}

@ARTICLE{014,
  author={Yang, Lei and Zhang, Xinyu and Yu, Jiaxin and Li, Jun and Zhao, Tong and Wang, Li and Huang, Yi and Zhang, Chuang and Wang, Hong and Li, Yiming},
  journal={IEEE Transactions on Intelligent Transportation Systems}, 
  title={MonoGAE: Roadside Monocular 3D Object Detection With Ground-Aware Embeddings}, 
  year={2024},
  volume={25},
  number={11},
  pages={17587-17601},
  doi={10.1109/TITS.2024.3412759}}

@ARTICLE{015,
  author={Gao, Honghao and Yu, Xinxin and Xu, Yueshen and Kim, Jung Yoon and Wang, Ye},
  journal={IEEE Transactions on Consumer Electronics}, 
  title={MonoLI: Precise Monocular 3-D Object Detection for Next-Generation Consumer Electronics for Autonomous Electric Vehicles}, 
  year={2024},
  volume={70},
  number={1},
  pages={3475-3486},
  doi={10.1109/TCE.2024.3353530}}

@inproceedings{016,
 author = {Jinrang, Jia and Li, Zhenjia and Shi, Yifeng},
 booktitle = {Advances in Neural Information Processing Systems (NeurIPS)},
 pages = {11703--11715},
 title = {MonoUNI: A Unified Vehicle and Infrastructure-side Monocular 3D Object Detection Network with Sufficient Depth Clues},
 volume = {36},
 year = {2023},
month={Dec.}
}

@inproceedings{017,
author = {Lee, Jeho and Jung, Chanyoung and Kim, Jiwon and Cha, Hojung},
title = {Panopticus: Omnidirectional 3D Object Detection on Resource-constrained Edge Devices},
year = {2024},
doi = {10.1145/3636534.3690688},
booktitle = {Proceedings of the 30th Annual International Conference on Mobile Computing and Networking (ACM)},
pages = {1207–1221},
numpages = {15},
}

@inproceedings{018,
 author = {Shen, Hangchi and Zheng, Qian and Wang, Huamin and Pan, Gang},
 booktitle = {Advances in Neural Information Processing Systems (NeurIPS)},
 pages = {92697--92720},
 title = {Rethinking the Membrane Dynamics and Optimization Objectives of Spiking Neural Networks},
 volume = {37},
 year = {2024},
month={Dec.}
}

@article{019,
title = {Analog Spiking U-Net integrating CBAM\&ViT for medical image segmentation},
journal = {Neural Networks},
volume = {181},
pages = {106765},
year = {2025},
issn = {0893-6080},
doi = {https://doi.org/10.1016/j.neunet.2024.106765},
author = {Yuqi Ma and Huamin Wang and Hangchi Shen and Shukai Duan and Shiping Wen}
}

@inproceedings{
020,
title={MonoDistill: Learning Spatial Features for Monocular 3D Object Detection},
author={Zhiyu Chong and Xinzhu Ma and Hong Zhang and Yuxin Yue and Haojie Li and Zhihui Wang and Wanli Ouyang},
booktitle={International Conference on Learning Representations (ICLR)},
year={2022},
month={Apr.}

}

@article{021,
  title={Monogrnet: A general framework for monocular 3d object detection},
  author={Qin, Zengyi and Wang, Jinglu and Lu, Yan},
  journal={IEEE transactions on pattern analysis and machine intelligence},
  volume={44},
  number={9},
  pages={5170--5184},
  year={2021},
  publisher={IEEE}
}

@inproceedings{022,
  author    = {Lis, K. and Kryjak, T. and Gorgoń, M.},
  title     = {LiFT: Lightweight, FPGA-tailored 3D object detection based on LiDAR data},
  booktitle = {International Workshop on Design and Architectures for Signal and Image Processing (DASIP)},
  year      = {2025},
  month     = {January},
  pages     = {28-40},

}

@article{023,
  author= {X. Zhang and H. Wang and H. Dong},
  title= {A Survey of Deep Learning-Driven 3D Object Detection: Sensor Modalities, Technical Architectures, and Applications},
  journal= {Sensors},
  year= {2025},
  volume= {25},
  number= {12},
  pages= {3668},
  month= {Dec.},
}

@article{027,
  title={Toward Energy-Efficient Spike-Based Deep Reinforcement Learning With Temporal Coding},
  author={Zhang, Malu and Wang, Shuai and Wu, Jibin and Wei, Wenjie and Zhang, Dehao and Zhou, Zijian and Wang, Siying and Zhang, Fan and Yang, Yang},
  journal={IEEE Computational Intelligence Magazine},
  volume={20},
  number={2},
  pages={45--57},
  year={2025},
}

@inproceedings{028,
  title={1.1 computing's energy problem (and what we can do about it)},
  author={Horowitz, Mark},
  booktitle={2014 IEEE international solid-state circuits conference digest of technical papers (ISSCC)},
  pages={10--14},
  year={2014},
}

@article{029,
  title={Toward Building Human-Like Sequential Memory Using Brain-Inspired Spiking Neural Models},
  author={Zhang, Malu and Luo, Xiaoling and Wu, Jibin and Belatreche, Ammar and Cai, Siqi and Yang, Yang and Li, Haizhou},
  journal={IEEE transactions on neural networks and learning systems},
  year={2025},
  volume={36},
  number={6},
  pages={10143-10155},
}

@article{030,
  title={Spike-Driven Lightweight Large Language Model With Evolutionary Computation},
  author={Zhang, Malu and Wei, Wenjie and Zhou, Zijian and Liu, Wanlong and Zhang, Jie and Belatreche, Ammar and Yang, Yang},
  journal={IEEE Transactions on Evolutionary Computation},
  year={2025},
  note = {doi: \url{https://doi.org/10.1109/TEVC.2025.3606613}}
 
}

@article{031,
  title={Event-driven learning for spiking neural networks},
  author={Wei, Wenjie and Zhang, Malu and Zhang, Jilin and Belatreche, Ammar and Wu, Jibin and Xu, Zijing and Qiu, Xuerui and Chen, Hong and Yang, Yang and Li, Haizhou},
  journal={arXiv preprint arXiv:2403.00270},
  year={2024}
}

@inproceedings{
032,
title={{QP}-{SNN}: Quantized and Pruned Spiking Neural Networks},
author={Wenjie Wei and Malu Zhang and Zijian Zhou and Ammar Belatreche and Yimeng Shan and Yu Liang and Honglin Cao and Jieyuan Zhang and Yang Yang},
booktitle={The Thirteenth International Conference on Learning Representations (ICLR)},
year={2025},
month={Apr.}

}

@article{033,
  title={Workload-balanced pruning for sparse spiking neural networks},
  author={Yin, Ruokai and Kim, Youngeun and Li, Yuhang and Moitra, Abhishek and Satpute, Nitin and Hambitzer, Anna and Panda, Priyadarshini},
  journal={IEEE Transactions on Emerging Topics in Computational Intelligence},
  volume={8},
  number={4},
  pages={2897--2907},
  year={2024},
}

@article{034,
  title={Modeling of Spiking Neural Network With Optimal Hidden Layer via Spatiotemporal Orthogonal Encoding for Patterns Recognition},
  author={Huang, Zenan and Chang, Yinghui and Wu, Weikang and Zhao, Chenhui and Luo, Hongyan and He, Shan and Guo, Donghui},
  journal={IEEE Transactions on Emerging Topics in Computational Intelligence},
  year={2025},
  volume={9},
  number={3},
  pages={2194-2207},
  publisher={IEEE}
}

@article{035,
  title={Efficient processing of spiking neural networks via task specialization},
  author={Lebdeh, Muath Abu and Yildirim, Kasim Sinan and Brunelli, Davide},
  journal={IEEE Transactions on Emerging Topics in Computational Intelligence},
  volume={8},
  number={5},
  pages={3603--3613},
  year={2024},
  publisher={IEEE}
}

@article{036,
  title={Spatio-temporal fusion spiking neural network for frame-based and event-based camera sensor fusion},
  author={Qiao, Guanchao and Ning, Ning and Zuo, Yue and Zhou, Pujun and Sun, Mingliang and Hu, Shaogang and Yu, Qi and Liu, Yang},
  journal={IEEE Transactions on Emerging Topics in Computational Intelligence},
  volume={8},
  number={3},
  pages={2446--2456},
  year={2024},
  publisher={IEEE}
}

\newpage
\vspace{-1.2cm}

\end{document}